\documentclass[pmlr,twocolumn]{jmlr} 

\usepackage[load-configurations=version-1]{siunitx} 
\usepackage{longtable}  
\usepackage[labelfont=bf]{caption}
\usepackage{float}
\usepackage{amssymb, amsfonts, bm}
\usepackage{mathtools}

\graphicspath{{./images/}}

\newcommand{\cD}{\mathcal{D}}

\newcommand{\cL}{\mathcal{L}}

\newcommand{\norm}[1]{\left\lVert#1\right\rVert}
\newcommand{\bffX}{{\bm{X}}}
\newcommand{\bffx}{{\bm{x}}}
\newcommand{\bffY}{{\bm{Y}}}
\newcommand{\bffy}{{\bm{y}}}

\newcommand{\bffZ}{{\bm{Z}}}

\newcommand{\bffB}{{\bm{B}}}

\newcommand{\bffA}{{\bm{A}}}

\newcommand{\bffs}{{\bm{s}}}
\newcommand{\bffW}{{\bm{W}}}

\newcommand{\bffD}{{\bm{D}}}

\newcommand{\bffP}{{\bm{P}}}

\newcommand{\bffV}{{\bm{V}}}

\theorembodyfont{\upshape}
\theoremheaderfont{\scshape}
\theorempostheader{:}
\theoremsep{\newline}

\jmlrvolume{LEAVE UNSET}
\jmlryear{2020}
\jmlrsubmitted{LEAVE UNSET}
\jmlrpublished{LEAVE UNSET}
\jmlrworkshop{Machine Learning for Health (ML4H) 2020} 

\title[Adversarially-Generated Autism Biomarkers]{Extended Abstract - Adversarial Factor Models for the\\Generation of Improved Autism Diagnostic Biomarkers}

\author{
\Name{William E. {Carson IV}} \Email{william.carson@duke.edu} \\
\addr{Department of Biomedical Engineering}
\AND
\Name{Dmitry {Isaev}} \\
\addr{Department of Biomedical Engineering}
\AND
\Name{Samantha {Major}} \\
\addr{Duke Center for Autism \& Brain Development}
\AND
\Name{Geraldine {Dawson}} \\
\addr{Duke Center for Autism \& Brain Development}
\AND
\Name{Guillermo {Sapiro}} \\
\addr{Department of Electrical \& Computer Engineering}
\AND
\Name{David E. {Carlson}} \\
\addr{Department of Civil and Environmental Engineering} \\
\addr{Duke University, Durham, NC 27708, U.S.}
}

\begin{document}

\maketitle

\begin{abstract}

Discovering reliable measures that inform on autism spectrum disorder (ASD) diagnosis is critical for providing appropriate and timely treatment for this neurodevelopmental disorder. In this work we present applications of adversarial linear factor models in the creation of improved biomarkers for ASD diagnosis. First, we demonstrate that an adversarial linear factor model can be used to remove confounding information from our biomarkers, ensuring that they contain only pertinent information on ASD. Second, we show this same model can be used to learn a disentangled representation of multimodal biomarkers that results in an increase in predictive performance. These results demonstrate that adversarial methods can address both biomarker confounds \textit{and} improve biomarker predictive performance.

\end{abstract}


\section{Introduction}
\label{sec:intro}

A major goal in autism spectrum disorder (ASD) research is to develop robust biomarkers that can improve diagnostic classification \citep{McPartland_2016}. Here, we explore adversarial methods for the generation of improved ASD diagnostic biomarkers, focusing specifically on linear factor models.

First, we use an adversarial linear factor model to address common confounds in ASD biomarkers. Intellectual disability (ID) is a common comorbidity present in approximately 40\% of individuals with autism \citep{centers2016prevalence}. Because of the overlap between ASD and ID, when studying the biological basis of autism it is often difficult to determine whether observed differences between typically-developing (TD) and ASD individuals can be attributed to ASD or differences in cognitive ability. Thus, methods to extract ASD-specific features \textit{independent} of information attributable to intellectual ability/disability are needed.

Finally, we demonstrate this adversarial model can improve biomarker diagnostic predictive performance. We use the adversarial model to maximize distinct information represented in different modalities by removing redundant information across modalities. The resulting disentangled multimodal biomarkers lead to a compressed representation as well as superior predictive performance.

\section{Methods}
\label{sec:methods}

\subsection{Data}

\subsubsection*{Cohort}

The cohort consisted of 31 children with ASD and 31 age matched typically-developing (TD) children. Simultaneously recorded EEG and videotaped behavior were collected while participants watched three video stimuli. Video content consisted of a woman singing nursery rhymes while gesturing, colorful dynamic toys that made noise, and bubbles silently cascading across the screen.

\subsubsection*{EEG Features}

Average power spectral density (PSD) from artifact-free one second epochs of EEG recording for each channel were binned into four power bands: theta (5-7 Hz), alpha (8-10 Hz), beta 1 (11-20 Hz), and beta 2 (21-30 Hz). PSD values were averaged over three scalp regions (frontal, central, and posterior). Twelve channels covering the left hemisphere, right hemisphere, and midline were included in each of the three regions. The final result was 12 PSD features for each of the three video stimuli, for a total of 36 PSD features (EEG features).

\subsubsection*{Behavior Features}

Behavior features were generated based on participant attention to video stimuli. Average Look Duration (ALD) \citep{isaev2020relative} for the three different video stimuli was calculated by dividing the total amount of time the participant was watching the screen when a video stimulus was presented by the number of looks at the screen during the video stimulus. Thus, each participant was assigned three ALD scores (behavior features), one for each of the video stimuli.

\begin{figure}[t]
\centering
\includegraphics[width=1.0\columnwidth]{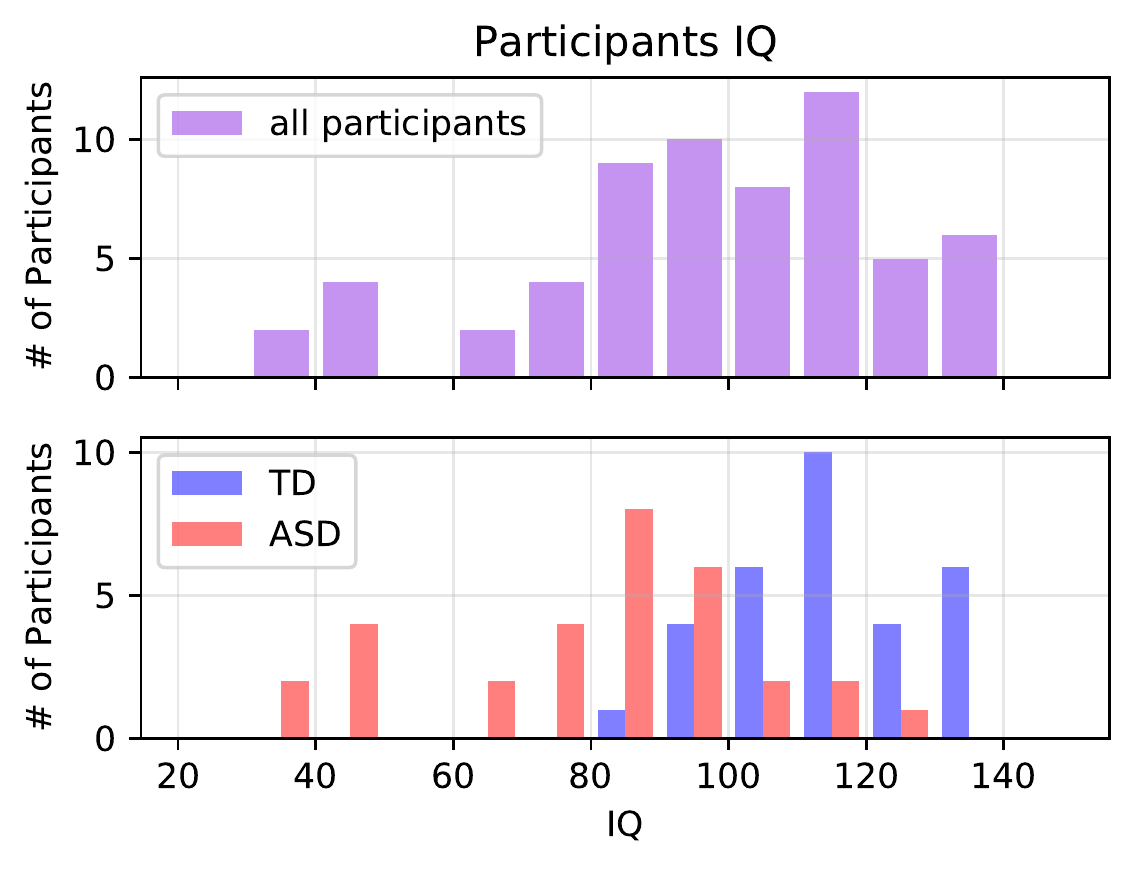}
\caption{The histogram of participant IQ depicts the issue of the IQ confound. The bi-modal distribution is a function of the disparate IQs of the TD and ASD groups.}
\label{fig:iq_hist}
\end{figure}

\begin{table*}[h]
\centering
\begin{tabular}{|l|c|c|}
\hline
\textbf{Behavior Features Preprocessing} & \textbf{AUC} (diagnosis) & \textbf{AUC} (IQ group) \\
\hline
none & 0.795 & 0.701 \\
\hline
PCA & 0.780 & 0.692 \\
\hline
adversarial PCA (IQ-invariant) & 0.631 & \textbf{0.500} \\
\hline
\end{tabular}
\caption{\label{tab:iq_correction} Classifier performance according to preprocessing method used. aPCA successfully removes excessive information on participant intelligence from behavior features.}
\end{table*}

\begin{table*}[t]
\centering
\begin{tabular}{|l|c|c|c|}
\hline
\textbf{Multimodal Features Preprocessing} & \textbf{\# of Features} & \textbf{AUC} (diagnosis) \\
\hline
none (na\"ive concatenation) & 39 & 0.680 \\
\hline
PCA & 23 & 0.787 \\
\hline
adversarial PCA & 19 & \textbf{0.813} \\
\hline
\end{tabular}
\caption{\label{tab:multimodal_results} Classifier performance according to multimodal feature set preprocessing. Disentangling multimodal features results in an increase in predictive performance.}
\end{table*}

\subsection{Adversarial Linear Factor Model}
\label{subsec:adv_model}

Factor models seek to explain the variance of observed data in terms of lower dimensional latent variables known as factors. The general formulation of a factor model can be defined as follows:
\begin{equation}\label{eq:factor_model}
    \min_{\{\bffs_i\}_{i=1,...,N},\Theta} \sum_{i=1}^N -\cL(\bffx_i|\Theta,\bffs_i),
\end{equation}
where $\{\bffx_i\}_{i=1:N}$ are the observed data, $\{\bffs_i\}_{i=1:N}$ are the factors, $\cL$ is the loss relating the factors to the data, and $\Theta$ represents the loadings - variables that relate the latent factors to the observed or primary data.

Here we describe a factor model that satisfies the following properties: ($i$) the learned representation contains information about the data of interest (primary data) and ($ii$) the learned representation does not contain information about concomitant data. To do this we reformulate the factor model as a minimax or adversarial objective:
\begin{equation}\label{eq:je_minimax}
\min_{A,\Theta}\max_\Phi \sum_{i=1}^N -\cL(\bffx_i,\bffy_i|\Theta,A)-\mu\cD(\bffy_i|A,\Phi)
\end{equation}
where $\{\bffs_i\}_{i=1:N}$ represents the concomitant data, $\cD$ is our adversarial loss, and $\Phi$ are the adversarial parameters. This objective balances fitting the observed data while making the factors unpredictive of the concomitant data. The degree of this unpredictability can be controlled through the adversarial strength tuning parameter $\mu$.

Note that the linear version of this model has an analytical solution computed by an augmented eigenspace, yielding efficient and reliable computation. Further methodology detailing the solutions to the analytic solution can be seen in Appendix \ref{app:app_a}. Intuitively, for a linear model at $\mu=0$ equation \ref{eq:je_minimax} is equivalent to Principal Component Analysis (PCA). Thus, moving forward this adversarial linear factor model will be referred to as adversarial PCA, or aPCA.

\subsection{Adversarially-Generated Biomarkers}

To address the common confound between intellectual ability and ASD, aPCA was used to remove information related to participant intelligence by making the features invariant to participant IQ. To check whether features contained excessive information on IQ we tested classifier ability to predict above or below average IQ (IQ = 100). Diagnostic ability of behavior features modified using aPCA was compared against that of behavior features with no preprocessing performed and behavior features decomposed using PCA.

\begin{figure}[t]
\centering
\includegraphics[width=1.0\columnwidth]{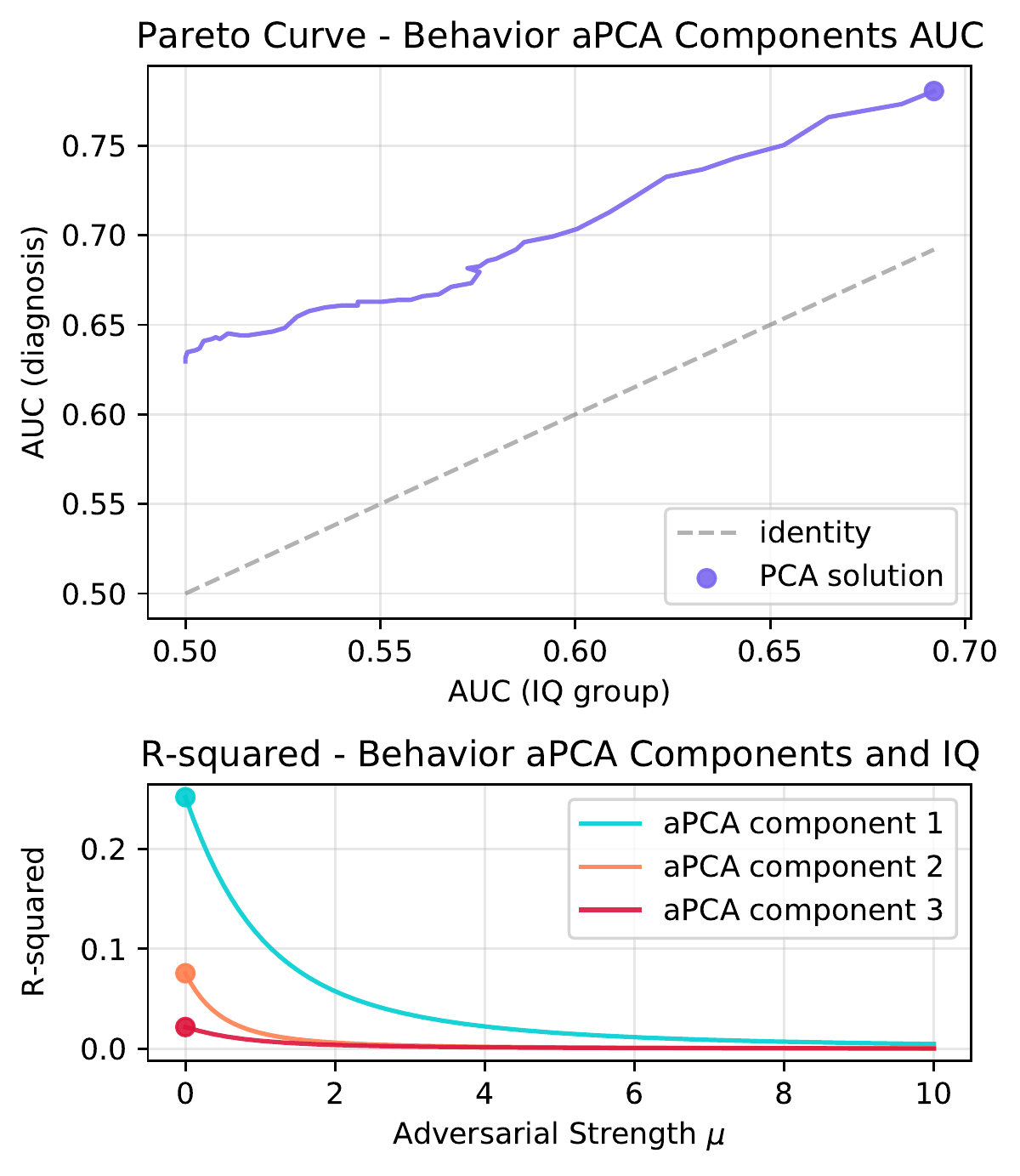}
\caption{Pareto curve of diagnosis and IQ predictability as adversarial strength is varied (top) and correlation between behavior aPCA components and IQ as a function of adversarial strength (bottom).}
\label{fig:iq_correction}
\end{figure}

\begin{figure}[t]
\centering
\includegraphics[width=1.0\columnwidth]{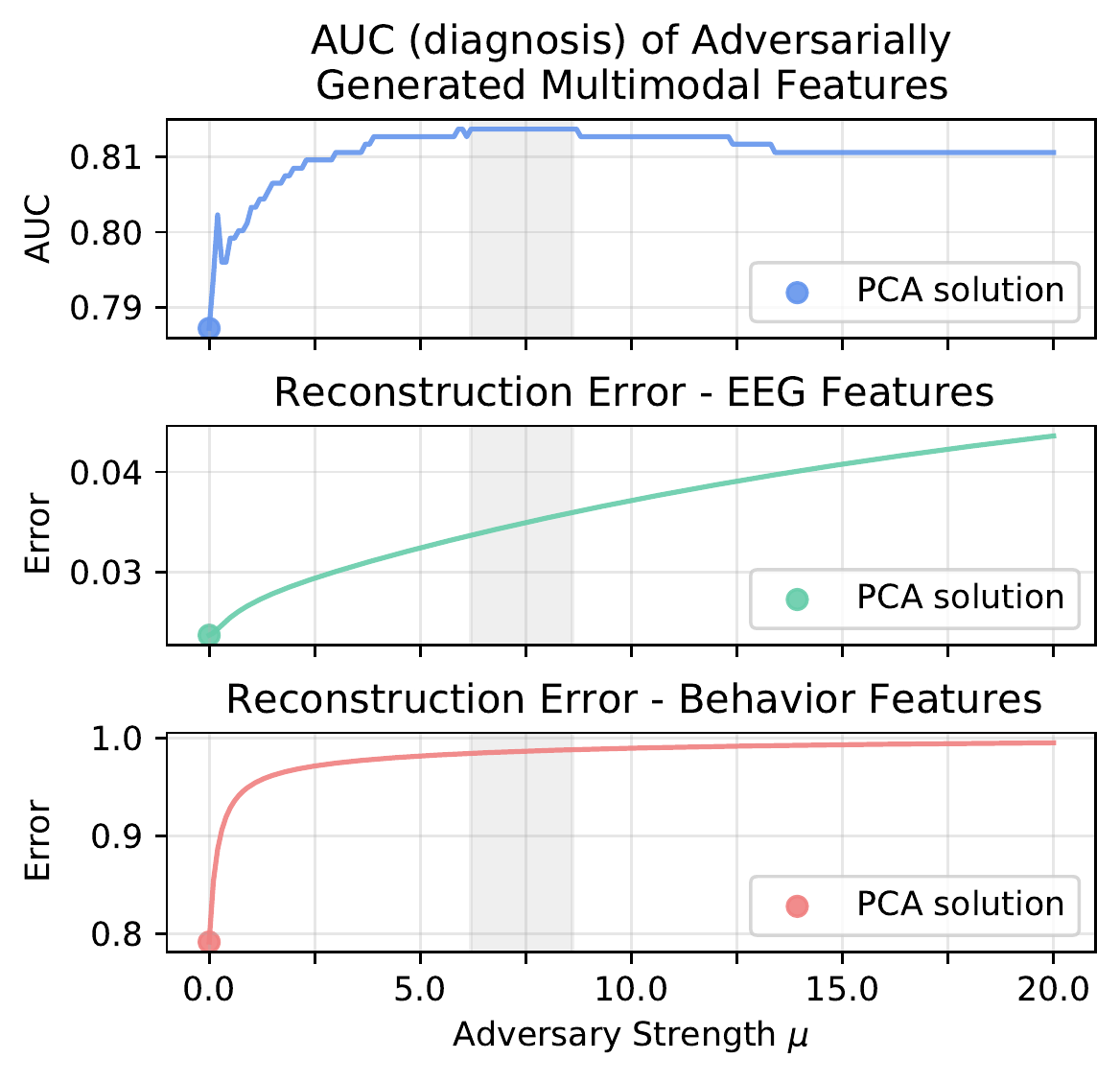}
\caption{Depicted are the AUC achieved using adversarial-generated multimodal features (top), the reconstruction error of EEG features (primary data) (middle), and behavior features (concomitant data) (bottom) as a function of adversarial strength. Shaded regions indicate adversarial strength values corresponding to max classifier performance.}
\label{fig:multimodal_auc}
\end{figure}

In the case of neural and behavioral features, neural features may contain redundant information on behavior as the brain informs behavior. Using aPCA, we create EEG features unpredictive of behavior, thus learning a disentangled multimodal representation. We compare diagnosis predictability of these disentangled features against diagnosis predictability of no feature preprocessing (na\"ive concatenation of features) and features decomposed using PCA.

Additional details on the biomarker evaluation process can be found in Appendix \ref{app:app_b}.

\section{Results and Discussion}

Results of using aPCA to enforce IQ-invariance in our features can be seen in Table \ref{tab:iq_correction}. Only behavior features are modified to be unpredictive of IQ as it was determined raw EEG features were not predictive of IQ (AUC = 0.504). aPCA is able to make behavior features completely unpredictive of IQ (AUC = 0.500) while still retaining information pertinent to ASD diagnosis (AUC = 0.631). The top plot in Figure \ref{fig:iq_correction} shows the tradeoff between diagnosis predictability and IQ predictability as adversarial strength is increased. The bottom plot in Figure \ref{fig:iq_correction} shows how the correlation between behavior aPCA components and IQ decreases as adversarial strength increases, thus demonstrating the effectiveness of the adversary.

Results of using aPCA to improve multimodal feature predictive performance is shown in Table \ref{tab:multimodal_results}. The learned disentangled representation of EEG and behavior features outperforms both na\"ive concatenation of features and PCA decomposition. The top plot in Figure \ref{fig:multimodal_auc} depicts classifier performance as a function of adversarial strength. The middle and bottom plots in Figure \ref{fig:multimodal_auc} show reconstruction error of the primary and concomitant data as a function of adversarial strength, respectively. The shaded regions correspond to max classifier performance where behavioral information has been sufficiently removed from EEG features but not so much so that useful information is lost. These results demonstrate that adversarial methods can successfully disentangle multimodal biomarkers, leading to improved predictive performance.







\newpage

\bibliography{main}


\appendix{}

\section{Analytical Solutions}
\label{app:app_a}

Analytical solution for the jointly encoded adversary. Here the linear form is given as 
\begin{equation}\label{eq:je_lin}
\min_{\bffW,\bffA}\max_{\bffD} \norm{\bffX-\bffW \bffA\bffZ}_F^2-\mu\norm{\bffY-\bffD \bffA\bffZ}_F^2
\end{equation}
where $\bffZ^T=[\bffX^T,\bffY^T]$. Equation \ref{eq:je_lin} can be equivalently represented as 
\begin{equation}\label{eq:je_lin2}
\min_{\bffW,\bffA}\max_{\bffD} \norm{\bffZ-\bffV \bffA\bffZ}_F^2-\mu^*\norm{\bffY-\bffD\bffA\bffZ}_F^2,
\end{equation}
where $\mu^*=\mu + 1$ and $\bffV=[\bffW^T,\bffD^T]^T$.

The solution to Equation (\ref{eq:je_lin2}) for $\bffV$ and $\bffD$ comes from scaled versions of the eigenvectors associated with the $l$ largest eigenvalues of the matrix
\[
\bffB = \begin{bmatrix}
\bffZ\bffZ^T & -\mu^* \bffZ\bffY^T \\
\bffY\bffZ^T & -\mu^* \bffY\bffP_{\bffZ}\bffY^T
\end{bmatrix}.
\]
with $e_i = [v_i^T d_i^T]$ where $e_i$ is the $i^{\text{th}}$ eigenvector and $v_i$ and $d_i$ are the associated loadings and discriminator and $\bffP_{\bffZ}=\bffZ^T(\bffZ\bffZ^T)^{-1}\bffZ$.

\section{Biomarker Evaluation}
\label{app:app_b}

Linear regression with a $l1$ penalty was the classifier used for all analyses. For all analyses, a parameter search was performed to over linear regression regularization parameter $C$ for values $C = 1\cdot10^{-2}, 1\cdot10^{-1}, ..., 1\cdot10^{6},$. In cases where dimensionality reduction was used (PCA and aPCA) a parameter search over the number of components $k$ was performed for values $k = 1, 2, ..., p$ where $p$ represents the original feature dimensionality. When using aPCA to decompose feature sets, a parameter search over the adversarial strength parameter $\mu$ for values $\mu = 0.0, 0.1, ..., 20.0$ was performed.

\end{document}